\documentclass[10pt,twocolumn,letterpaper]{article}

\usepackage[pagenumbers]{cvpr} 

\usepackage{kotex}

\usepackage{booktabs}
\usepackage{multirow}
\usepackage{multicol}
\usepackage{makecell}
\usepackage{setspace}

\def\MethodName{LayeringDiff}

\renewcommand{\paragraph}[1]{\vspace{2pt}\noindent\textbf{#1}~~}

\definecolor{cvprblue}{rgb}{0.21,0.49,0.74}
\usepackage[pagebackref,breaklinks,colorlinks,allcolors=cvprblue]{hyperref}

\title{LayeringDiff: Layered Image Synthesis via Generation, then  Disassembly\\with Generative Knowledge}

\author{Kyoungkook Kang$^{1}$ ~ ~ ~
Gyujin Sim$^{1}$ ~ ~ ~
Geonung Kim$^{2}$ ~ ~ ~\\
Donguk Kim$^{3}$ ~ ~ ~
Seungho Nam$^{3}$ ~ ~ ~
Sunghyun Cho$^{1,2}$\\[2mm]
POSTECH CSE$^{1}$ \& GSAI$^{2}$ \\
{\tt\small \{kkang831, sgj0402, k2woong92, s.cho\}@postech.ac.kr} \\
SHIFT UP$^{3}$ \\
{\tt\small \{dong, shnam48\}@shiftup.co.kr}
}

\spacing{0.93}
\begin{document}
\maketitle
\begin{abstract}
Layers have become indispensable tools for professional artists, allowing them to build a hierarchical structure that enables independent control over individual visual elements.
In this paper, we propose \MethodName{}, a novel pipeline for the synthesis of layered images, which begins by generating a composite image using an off-the-shelf image generative model, followed by disassembling the image into its constituent foreground and background layers.
By extracting layers from a composite image, rather than generating them from scratch, \MethodName{} bypasses the need for large-scale training to develop generative capabilities for individual layers.
Furthermore, by utilizing a pretrained off-the-shelf generative model, our method can produce diverse contents and object scales in synthesized layers. 
For effective layer decomposition, we adapt a large-scale pretrained generative prior to estimate foreground and background layers.
We also propose high-frequency alignment modules to refine the fine-details of the estimated layers. 
Our comprehensive experiments demonstrate that our approach effectively synthesizes layered images and supports various practical applications.
\end{abstract}
\vspace{-0.4cm}
   
\section{Introduction}
\label{sec:introduction}

Layers, featured in modern image editing tools such as Adobe Photoshop, are now indispensable components in professional image editing. 
When editing images, artists often create multiple layers, each containing a distinct visual element, and merge them to produce a complete image.
This hierarchical layered structure allows meticulous management and manipulation of each element without destroying the others, which allows artists to explore diverse compositions and effects, greatly expanding their creative potential.

Building upon recent advancements in image generative models~\cite{ldm,nonequilibriumthermodynamics,ddpm,ddpmbeatsgans,improvedddpm,scorebasedgenerativemodeling,dalle2,imagine,sdxl}, a few studies~\cite{text2layer,layerdiffuse,layerdiff} have proposed generating a layered image from a user prompt by fine-tuning a pretrained text-to-image generative model to generate foreground and background layers. 
However, fine-tuning a generative model requires vast collections of foreground and background layers, which are hard to collect.
To address this, they have also proposed synthetic dataset construction pipelines.
For example, Zhang~\etal~\cite{text2layer} generate foreground layers by extracting foreground objects based on salient object detection from a real-world image dataset, and then synthesize background layers by inpainting the holes where the foreground objects were extracted. 
Zhang and Agrawala~\cite{layerdiffuse} first synthesize foreground layers using a generative model trained on RGBA images sourced from the Internet. 
From the synthesized foreground layers, they produce background layers by outpainting the background regions and then inpainting the foreground regions after removing the foreground objects.

Despite the dataset synthesis pipelines of previous methods, training a layered image generative model remains costly. It requires substantial computational resources to generate a large volume of layered images and to fine-tune generative models. 
Besides, rigorous data filtering is necessary to remove low-quality images from synthesized datasets, which requires significant human labeling.
These pipelines may also introduce unwanted biases into fine-tuned models.
For instance, Text2Layer~\cite{text2layer} often generates low-quality images due to its training data synthesized by na\"ive saliency estimation and thresholding. 
Similarly, LayerDiffuse~\cite{layerdiffuse} tends to produce disproportionately large foreground objects that occupy most of the image area, as it learns the foreground distribution from an RGBA dataset, which is typically object-centric.

This paper proposes a novel pipeline for layered image synthesis, named \emph{\MethodName{}}.
\MethodName{} first generates a composite image using an off-the-shelf image generative model, and then disassembles it into its constituent foreground and background layers.
This two-step approach offers a couple of distinct advantages. 
Firstly, by reframing layered image synthesis as a layer decomposition problem from a composite image, we can effectively avoid the need for a large-scale training dataset. Specifically, our initial step leverages an off-the-shelf image generative model without requiring fine-tuning. Furthermore, layer decomposition is significantly easier than synthesis and can be effectively achieved with a small amount of training data.
Secondly, leveraging a pretrained off-the-shelf generative model, \MethodName{} can synthesize layered images with a wide range of content and object scales, while also supporting the integration of various off-the-shelf models, such as ControlNet, that accommodate different conditions.

\MethodName{} operates in three stages. Firstly, the initial image generation stage generates an initial composite image using an off-the-shelf generative model. Next, the foreground determination stage identifies the foreground area based on the input text prompt. Lastly, the layering stage separates the image into foreground and background layers, which are then re-combined to produce the final composite image.
Among these stages, the layering stage is the most crucial. This stage leverages a generative prior to effectively decompose an input image into its constituent layers. To this end, we introduce a Foreground and Background Diffusion Decomposition (FBDD) module.
To further enhance the high-frequency details in decomposed layers, we also introduce a high-frequency alignment (HFA) module.

Our extensive experiments show that \MethodName{} outperforms existing methods, providing more diverse and natural foreground and background layers, making it highly practical for a wide range of applications.
Our contributions are summarized as follows:
\begin{itemize}
    \item We propose \MethodName{}, an effective pipeline for high-quality layered image synthesis without the need for large-scale training, achieved by reframing the task as a layer decomposition problem.
    \item For effective layer decomposition, we propose adapting a powerful generative prior to estimate both foreground and background layers, with the proposed FBDD module and HFA module.
    \item We demonstrate that our approach outperforms existing layered image synthesis approaches through extensive experiments including a user study. We also showcase diverse practical applications.
\end{itemize}

\section{Related Work}
\label{sec:relatedwork}

\subsection{Text-based Layered Image Synthesis}

Layered image synthesis has gained increasing attention, sparked by its practical potential~\cite{text2live,text2layer,layerdiffuse,layerdiff}.   
Recent approaches focus on fine-tuning text-to-image generative models to generate layered images based on user prompts~\cite{text2layer,layerdiffuse,layerdiff}.
Zhang~\etal~\cite{text2layer} introduce an autoencoder to embed both foreground and background layers within a unified latent representation. They then train a diffusion model on these latent representations to capture the joint distribution of both layers.
Zhang and Agrawala~\cite{layerdiffuse} first train a base generative model to generate a foreground layer with transparency using RGBA images. To synthesize layered images, they extend the base model to produce both foreground and background layers, employing two LoRAs~\cite{lora} with shared attention to facilitate coordinated synthesis across layers.
Huang~\etal~\cite{layerdiff}, aiming at multi-layered image synthesis, propose a 3D diffusion UNet that jointly denoises multiple random noises into distinct layers, which together form a composite image.

However, these network fine-tuning methods are constrained by the quality and diversity of their training data and require large-scale training to achieve high-quality models.
In this paper, we overcome these challenges by focusing on decomposing each layer from a composite image generated using a pretrained high-quality generative model, rather than generating them from scratch.

\subsection{Image Matting and Inpainting}
Layer decomposition can be considered an image matting task. Recent learning-based matting approaches primarily focus on the accurate estimation of alpha mattes, which represent the transparency of a foreground layer~\cite{vitmatte,ppmatting,fbalphamatting,deepimagematting,contextawareimagematting,mattinganything,diffmatte,matteformer}. However, the estimation of pixel values for foreground and background layers has been relatively overlooked.
To estimate the pixel values of layers, a typical approach is to estimate an alpha matte from an input image using neural networks, then use optimization-based methods that rely on simple priors such as local smoothness and color linearity~\cite{closedformimagematting,fastcolor}.
Unfortunately, this approach is limited due to its simple priors, especially in handling large occluded regions in background layers. 
Recently, a few learning-based methods have been proposed to directly estimate both the alpha mattes and pixel values~\cite{samplenet,fbalphamatting,contextawareimagematting}. However, these approaches also fail to handle large occluded regions due to their regression-based networks, which lack the ability to synthesize new content.

One potential option for handling large occluded regions in background layers is to employ inpainting techniques~\cite{ldm,lamainpainting}.
However, existing inpainting techniques rely on binary masks to indicate image regions to be inpainted, thus cannot properly handle regions occluded by semi-transparent foreground objects.
Our layer decomposition approach tackles the aforementioned limitations of previous matting and inpainting techniques by leveraging generative priors and the HFA module.

\begin{figure*}
    \centering
    \includegraphics[width=1\linewidth]{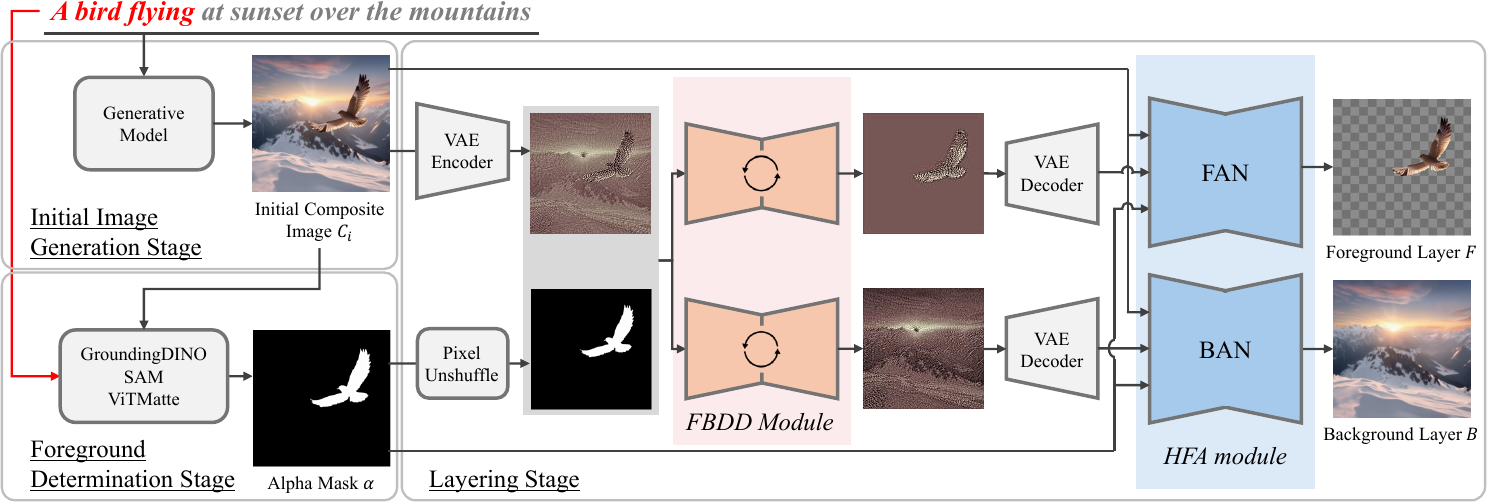}
    \vspace{-0.6cm}
    \caption{Overview of \MethodName{}. From an input text prompt $T$ including foreground prompt $T_F$ (red words), initial image generation stage synthesizes an initial composite image $C_i$.
    Then, foreground determination stage identifies a foreground region based on the foreground prompt $T_F$ and produce an alpha mask $\alpha$.
    Lastly, layering stage separates $C_i$ into a foreground layer $F$ and a background layer $B$.
    }
    \label{fig:1_1_pipeline}
    \vspace{-0.4cm}
\end{figure*}

\section{\MethodName{}}
\cref{fig:1_1_pipeline} illustrates the overview of \MethodName{}.
\MethodName{} starts with an input text prompt $T$ that describes a composite image and a set of indices $\mathbb{I}_F$ pointing to the words in the input text prompt corresponding to the foreground layer. For instance, for the input text prompt ``A bird flying at sunset over the mountains,'' $\mathbb{I}_F$ can be defined as $\mathbb{I}_F=\{0,1,2\}$ to indicate the sub-prompt ``A bird flying.'', which we refer to as the foreground prompt $T_F$.

Given $T$ and $\mathbb{I}_F$, \MethodName{} generates a foreground layer $F$ corresponding to $T_F$, a background layer $B$, and an alpha mask $\alpha$ so that their combination constructs a natural-looking composite image $C$ that reflects $T$.
Specifically, $C$ can be modeled using an image composition model, as:
\begin{equation}
    C = \alpha\cdot F + (1-\alpha)\cdot B .
    \label{eq:composition}
\end{equation}
To this end, \MethodName{} performs initial image generation, foreground determination, and layering stages to produce an output layered image.
In the following subsections, each stage is explained in detail.

\subsection{Initial Image Generation Stage}
From an input prompt $T$, the initial image generation stage synthesizes an initial composite image $C_i$, which will be converted to a layered representation in the following stages, using an off-the-shelf image generative model.
In this stage, any text-conditioned generative model can be employed such as a standard text-to-image model and a layout-conditioned model such as ControlNet~\cite{controlnet} to support additional input modalities such as an edge map and a depth map.
In our experiments, we adopt Stable Diffusion XL~\cite{sdxl} as the default generative model.

\subsection{Foreground Determination Stage}

The foreground determination stage identifies a foreground region based on the foreground prompt $T_F$, and generates an alpha mask to represent its boundary and transparency.
To this end, we adopt the automatic alpha mask estimation pipeline proposed in MatteAnything~\cite{mattinganything}.
First, a foreground bounding box is detected in the initial composite image $C_i$ based on the foreground prompt $T_F$ using Grounding DINO~\cite{groundingdino}, an open-vocabulary object detection model.
Next, a foreground semantic mask is estimated from the bounding box using the semantic segmentation model, Segment-Anything-Model (SAM)~\cite{sam}.
From the semantic mask, a trimap is generated by applying morphological dilation and erosion operations on the mask and assigning a value of 1 (foreground) to the eroded mask, 0 (background) to the outside of dilated mask, and 0.5 (unknown) to the region between the dilated and eroded boundaries.

The trimap is further refined by detecting transparent areas in the initial composite image $C_i$ using Grounding DINO~\cite{groundingdino} and assigning a value of 0.5 to the areas where the foreground region intersects with the transparent areas.
Finally, the predicted trimap and the initial composite image $C_i$ are passed to a matting model to estimate the alpha mask $\alpha$ of the foreground layer. 
In our experiments, we adopt ViTMatte~\cite{vitmatte} for matting due to its high accuracy.
For further details on the automatic alpha mask estimation pipeline, refer to MatteAnything~\cite{mattinganything}.

\subsection{Layering Stage}

Given an initial composite image $C_i$ and a foreground alpha mask $\alpha$, the layering stage decomposes $C_i$ into a foreground layer $F$ and a background layer $B$, based on the image composition model in \cref{eq:composition}.
Although $\alpha$ is given, layer decomposition is a severely ill-posed problem involving the estimation of six unknown variables (RGB values of foreground and background layers) with only four known values (RGB values of a composite image and an alpha) per pixel. The challenge is further compounded when the foreground layer $F$ is nearly opaque ($\alpha \approx 1$), concealing any information about the background layer $B$.

To synthesize high-quality natural-looking layers and overcome the aforementioned ill-posedness, the layering stage adopts an FBDD module, which is based on the latent diffusion model (LDM)~\cite{ldm}, and an HFA module.
Specifically, in the layering stage, we first encode the initial composite image $C_i$ into the latent representation via the encoder of LDM's Variational Autoencoder (VAE).
We also resize the alpha mask $\alpha$ to match the spatial resolution of the latent space.
We employ the pixel unshuffle operator~\cite{pixelunshuffle} to preserve information while resizing the alpha mask.
Subsequently, the FBDD module synthesizes foreground and background layers in the latent space, using the encoded $C_i$ and resized $\alpha$ as conditions.
The synthesized images are then decoded by the VAE decoder, producing intermediate foreground and background layers, $\hat{F}$ and $\hat{B}$. The HFA module enhances the high-frequency details in $\hat{F}$ and $\hat{B}$, generating the outputs $F$ and $B$.
The final $F$ and $B$ construct the final composite image $C$.

Both the FBDD and HFA modules are effectively trained using a much smaller dataset composed of synthetic composite images, as their task is relatively simpler than the comprehensive layer synthesis required by previous methods.
In reality, while previous methods typically use millions of training samples, we only use 20,000 training samples for the FBDD and HFA modules. 
In the following, we provide further details on the FBDD and HFA modules.

\paragraph{FBDD module}
For effective layer decomposition, the FBDD module leverages a generative prior pretrained on a large-scale dataset.
To this end, the FBDD module consists of two diffusion models: one for the foreground layer and another for the background layer.
Specifically, the FBDD module employs the diffusion UNet of the LDM~\cite{ldm}. 
Each of the UNets in both models takes an input consisting of a channel-wise concatenation of an intermediate latent image that starts with random noise, the latent representation of an initial composite image, and a resized alpha mask, and produces a denoised latent image corresponding to each layer.
The UNets are repeatedly performed to obtain resulting decomposed layers, following the iterative process of LDM.

Among various LDM-based models, we use the network architecture and pretrained weights of the Stable Diffusion 2 inpainting model for our VAE and diffusion UNets\footnote{https://huggingface.co/stabilityai/stable-diffusion-2-inpainting} due to its high image generation quality and task similarity. While the Stable Diffusion 2 inpainting model typically requires a text prompt, we fine-tune it to use a null prompt for both foreground and background layers. For more details on the implementation, refer to the supplementary material.

\begin{figure}
    \centering
    \includegraphics[width=\linewidth]{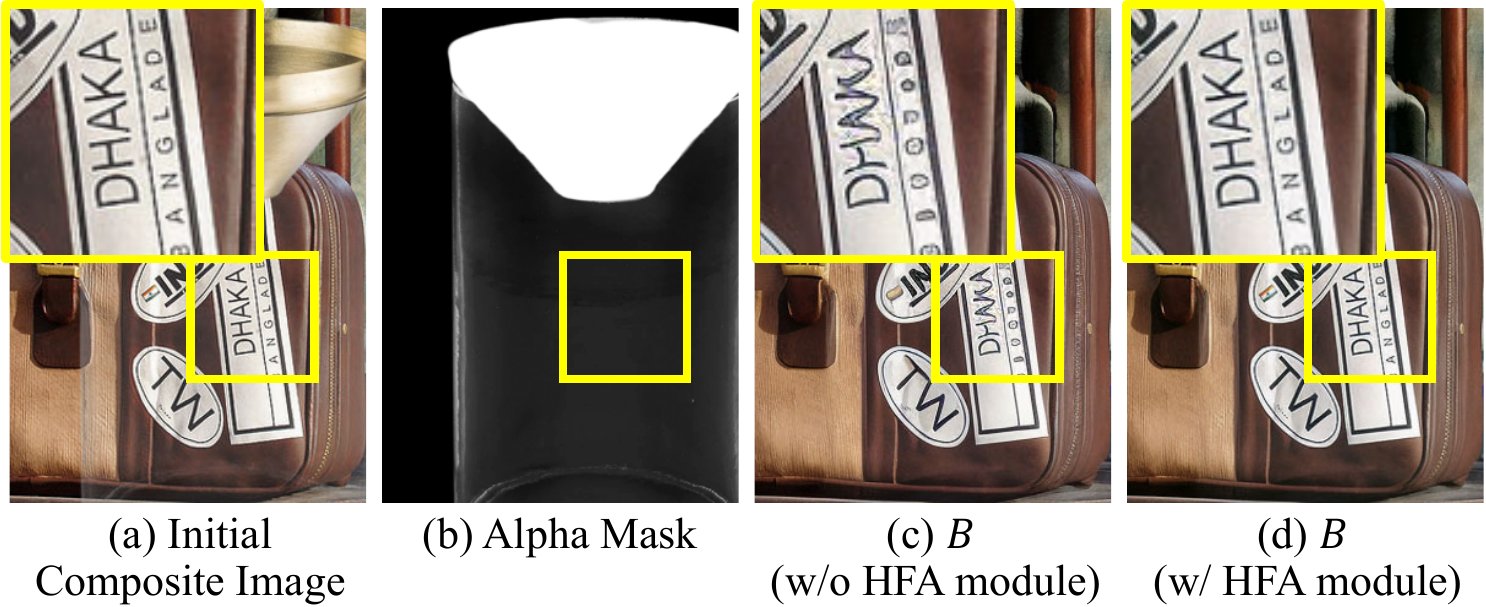}
    \vspace{-0.6cm}
    \caption{
    Decomposed layers by the FBDD module may suffer from degraded texture quality (c). HFA module enhance high-frequency details in these layers (d) using those from the initial composite image (a). Note that the text in the background is covered by the semi-transparent plane in the foreground layer in (a).
    }
    \label{fig:why_alignment_module_need}
    \vspace{-0.6cm}
\end{figure}

\paragraph{HFA module}
While the FBDD module can effectively decompose an initial composite image, decomposed layers may suffer from degraded texture qualities as shown in \cref{fig:why_alignment_module_need}.
This is mainly due to the fundamental difficulty of the layer decomposition task, which needs to create new contents while preserving existing information.
To enhance textures degraded by the FBDD module, the HFA module fuses the high-quality texture in the initial composite image into the output of the FBDD module.

Specifically, the HFA module consists of two sub-networks: the foreground alignment network (FAN), and the background alignment network (BAN).
Each network takes an initial composite image $C_i$, an alpha mask $\alpha$, and a decoded layer $\hat{F}$ or $\hat{B}$ from the FBDD module as input, and produces a refined foreground layer $F$ or a refined background layer $B$, respectively.
We adopt the UNet architecture for FAN and BAN, and initialize them with random weights.
Further details on the network architectures are provided in the supplementary material.
After FAN and BAN, we further refine $F$ and $B$ by directly copying pixel values from $C_i$ for the completely visible regions corresponding to $\alpha=1$ and $\alpha=0$, respectively.

\subsection{Training of \MethodName{}}
\label{subsec:training}
\begin{figure}
    \centering
    \includegraphics[width=\linewidth]{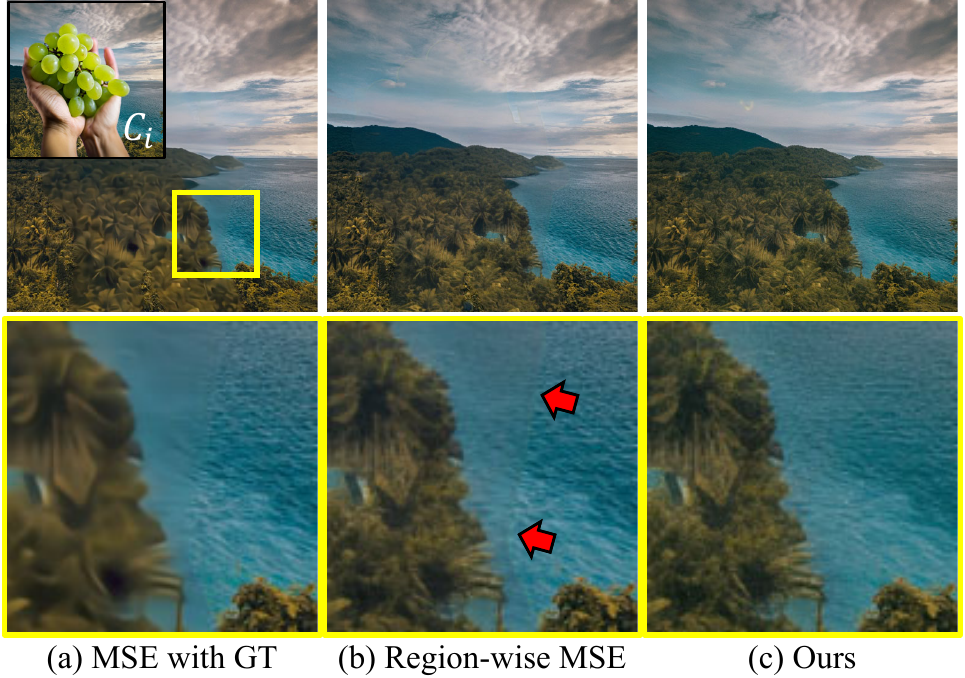}
    \vspace{-0.6cm}
    \caption{
    Qualitative comparison of backgrounds $B$ produced by BANs trained using different loss functions.
    The inset in the top-left image represents input composite image.
    }
    \label{fig:1_3_ablation}
    \vspace{-0.5cm}
\end{figure}

In our framework, we train only the FBDD and HFA modules, while keeping the other components, e.g., the initial image generation model, and the VAE encoder and decoder, fixed to their pretrained weights.
The FBDD and HFA modules are trained independently using a synthetically generated composite image dataset.
For the FBDD module, we fine-tune a pretrained LDM using the v-prediction loss~\cite{vprediction}.
In the following, we introduce our training dataset construction process, and the training strategies for the two sub-networks, FAN and BAN, of the HFA module.

\paragraph{Dataset construction}
We adopt a composite image synthesis strategy commonly adopted by image matting approaches~\cite{deepimagematting}, where composite images are generated by randomly pairing foreground and background images from their respective datasets and combining them using \cref{eq:composition}. 
For the foreground dataset, we use the MAGICK dataset~\cite{magick}, which provides 150K RGBA foreground images.
For the background dataset, we use the BG-20k dataset~\cite{bg20k}, which consists of 20K background images with no salient objects.
We perform bicubic interpolation to resize the composite images so that their shorter side is of 512 pixels, and crop the center region of size $512\times512$.

\paragraph{Training of the HFA module}
We train FAN and BAN using different loss functions to account for the different characteristics of the foreground and background layers, i.e., while the initial composite image provides all necessary information for the foreground layer, it lacks information for occluded regions in the background layer.
To train FAN, we employ a simple MSE loss to minimize the pixel-level discrepancies between the final output $F$ of FAN and the ground-truth foreground image $F_{gt}$.
However, training BAN using a simple MSE loss leads to low-quality results with artifacts such as seams between occluded and non-occluded regions, and false high-frequency textures in occluded regions, due to the aforementioned characteristic of the background layer, as shown in \cref{fig:1_3_ablation} (a).

To train BAN, we take into account the different regions within the background layer. Specifically, for completely or partially visible regions where $\alpha<1$, we can utilize information from the initial composite image to refine the texture of the background layer. Conversely, for completely occluded regions where $\alpha=1$, we can only use information of the background layer from the FBDD module, $\hat{B}$, as there is no available information in the initial composite image.
However, $\hat{B}$ may exhibit not only different textures but also slightly deviated colors from those of the ground-truth background layer $B_{gt}$, due to the diffusion process in the FBDD module.
Not considering such discrepancies in colors during the training of BAN may result in visible seams between different regions, as shown in \cref{fig:1_3_ablation} (b) where BAN is trained with a region-wise MSE loss, which minimizes differences between $B$ and $B_{gt}$ in visible regions and between $B$ and $\hat{B}$ in occluded regions.

Taking into account all the aforementioned aspects, we propose a loss function $\mathcal{L}_\textrm{BAN}$ for training BAN, defined as:
\begin{equation}
    \mathcal{L}_\textrm{BAN}=\mathcal{L}_\textrm{MSE}(B,B_\textrm{gt}) + \lambda \mathcal{L}_\textrm{H}(B,\hat{B}),
    \label{eq:L_BAN}
\end{equation}
where the first term on the right-hand side is an MSE loss that promotes $B$ to have pixel values close to the ground-truth $B_\textrm{gt}$.
The second term is a high-frequency error loss that encourages the final output $B$ to follow the high-frequency details of the FBDD output $\hat{B}$, whose definition will be given later.
$\lambda$ is a weight for the high-frequency error loss, which is set to $0.2$ in our implementation.

We uniformly apply both losses across the background layer.
Nevertheless, BAN is trained to handle different regions in the background layer effectively.
For visible regions ($\alpha<1$), BAN is trained to exploit information from $C_i$ to minimize $\mathcal{L}_\textrm{MSE}$, since $\mathcal{L}_\textrm{MSE}$ is dominant as we set $\lambda$ low.
For completely occluded regions ($\alpha=1$), BAN cannot be trained to use $C_i$ to minimize $\mathcal{L}_\textrm{MSE}$.
Instead, the effect of the high-frequency error loss $\mathcal{L}_\textrm{H}$ kicks in, aligning the final output $B$ with the textures of the FBDD output $\hat{B}$.
Additionally, constraining the high-frequency components using $\mathcal{L}_\textrm{H}$ ensures that $B$ matches the colors of $B_\textrm{gt}$ without seams between regions, shown in \cref{fig:1_3_ablation} (c).

We define the high-frequency error loss $\mathcal{L}_\textrm{H}$ using the Haar wavelet transform, which is a multi-scale frequency decomposition technique. Specifically, $\mathcal{L}_\textrm{H}$ is defined as:
\begin{equation}
    \mathcal{L}_\textrm{H}(B,\hat{B})=\sum_{s}\sum_{k}\frac{1}{N_s}\| \mathcal{H}_{s,k} (B) - \mathcal{H}_{s,k} (\hat{B})\|^2
\end{equation}
where $\mathcal{H}_{s,k}$ is a Haar function of scale $s$ and direction $k$ with $k \in \{\textrm{horizontal}, \textrm{vertical}, \textrm{diagonal}\}$. We use $s\in\{0,1,2\}$ in our experiments. $N_s$ is the number of pixels in scale $s$.

\begin{figure*}[t!]
    \centering
    \includegraphics[width=\linewidth]{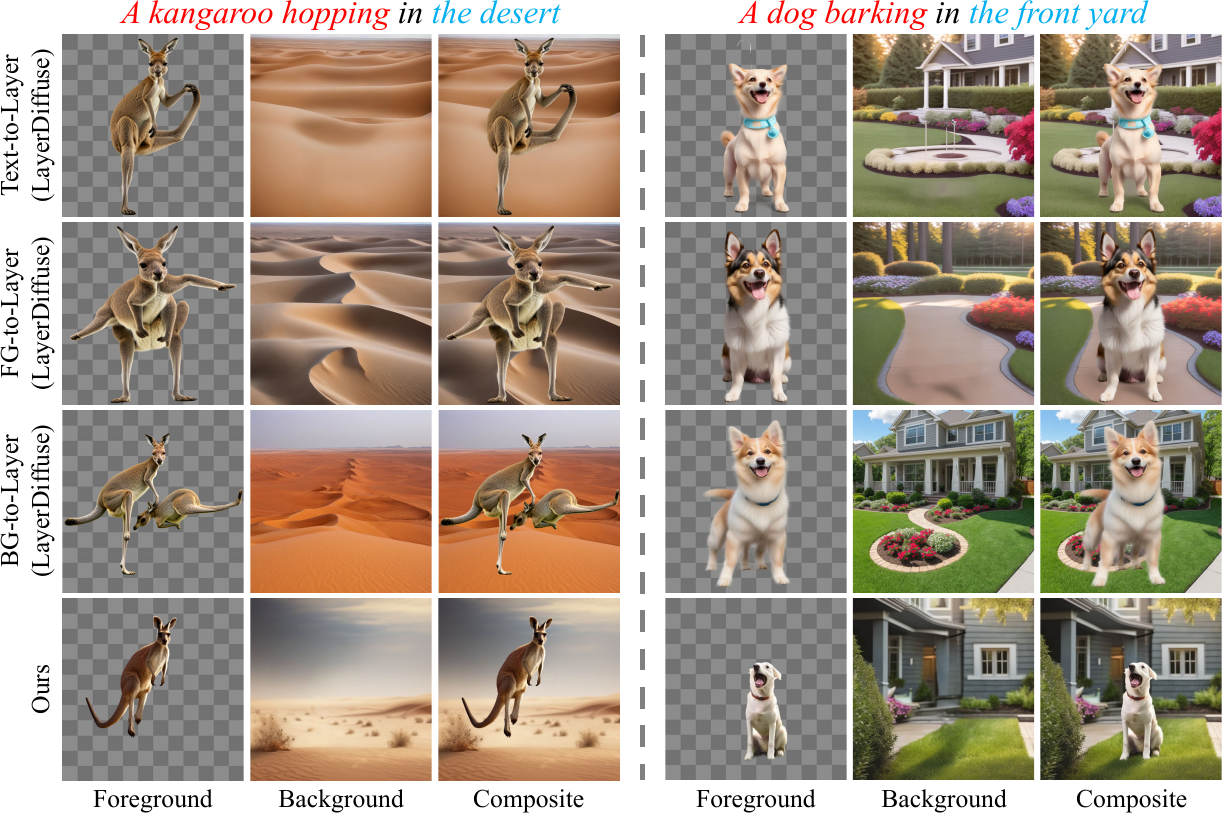}
    \vspace{-0.4cm}
    \caption{
    Qualitative comparison of layered images generated by the three models of LayerDiffuse~\cite{layerdiffuse} and our method for input prompts, positioned at the top of each example. 
    In each prompt, the red words denote the foreground prompt, while the blue words represents the background prompt.
    LayerDiffuse models tends to produce foreground objects disproportionately large relative to the background, whereas our method generates realistic, well-proportioned layered images.
    }
    \label{fig:5_1_qualitative_comparison}
\end{figure*}
\begin{figure}
    \centering
    \includegraphics[width=\linewidth]{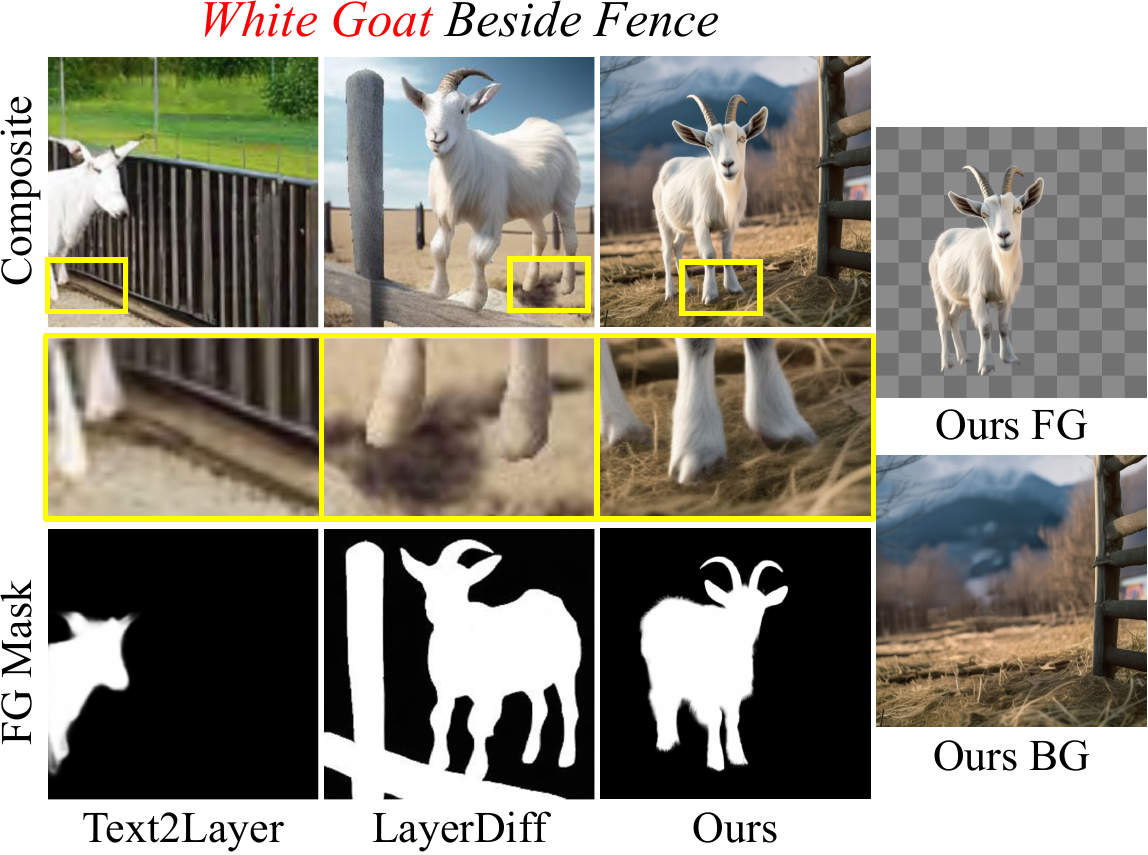}
    \vspace{-0.6cm}
    \caption{
    Qualitative comparison of layered images generated by Text2Layer~\cite{text2layer}, LayerDiff~\cite{layerdiff}, and our method for an input prompt, positioned at the top of the figure.
    }
    \label{fig:5_1_qualitative_comparison2}
    \vspace{-0.6cm}
\end{figure}

\begin{table*}[t]
\begin{center}
\caption{Quantitative evaluation of generated foreground, background, and final composite layers produced by the three models of LayerDiffuse~\cite{layerdiffuse} and our method. 
}
\vspace{-0.2cm}
\label{tab:tbl_1_main_quantitative_evaluation}
\scalebox{0.86}{
\begin{tabular}{c|cccc|cccc|ccc}
\Xhline{3\arrayrulewidth}

& \multicolumn{4}{c|}{Foreground} 
& \multicolumn{4}{c|}{Background} 
& \multicolumn{3}{c}{Composite} \\
\hline

Metrics
& FID$\downarrow$ & KID$\downarrow$ & CLIP score$\uparrow$ & FG MIoU$\uparrow$
& FID$\downarrow$ & KID$\downarrow$ & CLIP score$\uparrow$ & FG MIoU$\downarrow$
& FID$\downarrow$ & KID$\downarrow$ & CLIP score$\uparrow$  \\
\hline 

T2L (LD) & \textbf{127.14} & 0.033          & 28.72          & 0.62 
         & 175.58          & 0.051          & 27.87          & 0.25 
         & 134.51          & 0.021          & 30.10 \\

F2L (LD) & 146.15          & \textbf{0.032} & 28.43          & 0.72 
         & 180.13          & 0.050          & \textbf{28.15} & 0.24 
         & 143.50          & 0.023          & 29.57 \\

B2L (LD) & 128.20          & \textbf{0.032} & 28.62          & 0.65 
         & 207.39          & 0.038          & 27.70          & 0.22 
         & 134.57          & 0.018          & 30.05 \\

\rowcolor[HTML]{FFF9C4}  
Ours     & 133.76          & 0.037          & \textbf{28.86} & \textbf{0.87} 
         & \textbf{138.45} & \textbf{0.025} & 26.72          & \textbf{0.14}
         & \textbf{121.05} & \textbf{0.014} & \textbf{30.74} \\
\Xhline{3\arrayrulewidth}
\end{tabular}}
\end{center}
\vspace{-0.3cm}
\end{table*}
\begin{table}[t]
\begin{center}
\caption{
Analysis of synthesized foreground layers by each method, comparing mean and standard deviation on four metrics.
}
\vspace{-0.2cm}
\label{tab:tbl_2_foreground_layer_analysis}
\scalebox{0.80}{
\begin{tabular}{c|c|cccc}

\Xhline{3\arrayrulewidth}
  &   & LD (T2L) & LD (F2L) & LD (B2L) & \cellcolor[HTML]{FFF9C4} Ours \\
\hline

Occupancy  & $\mu$     & 37.88 & 35.32 & 32.89 & \cellcolor[HTML]{FFF9C4}23.06 \\
Ratio      & $\sigma$  & 17.57 & 14.90 & 15.62 & \cellcolor[HTML]{FFF9C4}17.50 \\
\hline

Longest    & $\mu$     & 95.58 & 95.06 & 92.27 & \cellcolor[HTML]{FFF9C4}69.50 \\
Span       & $\sigma$  & \phantom{0}6.41 & \phantom{0}6.34 & 10.93 & \cellcolor[HTML]{FFF9C4}21.69 \\
\hline

Vertical   & $\mu$     & 51.12 & 51.07 & 51.78 & \cellcolor[HTML]{FFF9C4}49.18 \\
Center     & $\sigma$  & \phantom{0}3.75 & \phantom{0}2.59 & \phantom{0}4.50 & \cellcolor[HTML]{FFF9C4} \phantom{0}8.82 \\
\hline

Horizontal & $\mu$     & 49.37 & 49.76 & 49.66 & \cellcolor[HTML]{FFF9C4}49.07 \\
Center     & $\sigma$  &  \phantom{0}2.60 &  \phantom{0}3.10 &  \phantom{0}2.64 & \cellcolor[HTML]{FFF9C4} \phantom{0}9.48 \\
\Xhline{3\arrayrulewidth}

\end{tabular}
}
\end{center}
\vspace{-0.6cm}
\end{table}
\begin{table}[t]
\begin{center}
\caption{
User study results, which reports average scores for two factors, text alignment and image quality, rated on a scale of 1 (poor) to 5 (excellent) by 24 participants.
}
\vspace{-0.2cm}
\label{tab:tbl_3_userstudy}
\scalebox{0.80}{
\begin{tabular}{c|c|ccc>{\columncolor[HTML]{FFF9C4}}c}

\Xhline{3\arrayrulewidth}
  &   & LD (T2L) & LD (F2L) & LD (B2L) & Ours \\
\hline

\multirow{2}{*}{FG}    & Text Align. $\uparrow$ & 3.86 & 3.67 & 3.55 & \textbf{4.36} \\
                       & Image Qual. $\uparrow$ & 3.53 & 3.38 & 2.91 & \textbf{4.12} \\
\hline

\multirow{2}{*}{BG}    & Text Align. $\uparrow$ & 4.18 & 3.91 & 4.33 & \textbf{4.47} \\
                       & Image Qual. $\uparrow$ & 3.61 & 3.25 & 4.27 & \textbf{4.33} \\
\hline

\multirow{2}{*}{Comp.} & Text Align. $\uparrow$ & 3.61 & 3.12 & 3.00 & \textbf{4.34} \\
                       & Image Qual. $\uparrow$ & 2.97 & 2.42 & 1.85 & \textbf{4.07} \\
\Xhline{3\arrayrulewidth}

\end{tabular}
}
\end{center}
\vspace{-0.6cm}
\end{table}

\section{Experiments}
\label{sec:experiments}
\subsection{Comparative Evaluation}

\paragraph{Baselines}
We compare the quality of layered images generated by our approach primarily with those of LayerDiffuse~\cite{layerdiffuse}, as it is the state-of-the-art layered image synthesis method, and the only method that provides its source code. 
We also include a qualitative comparison with Text2Layer~\cite{text2layer} and LayerDiff~\cite{layerdiff} using examples from their respective papers.
LayerDiffuse offers three models: text-to-layer (T2L), foreground-to-layer (F2L), and background-to-layer (B2L). The T2L model uses foreground and background prompts as input, the F2L model uses a foreground layer and a background prompt, and the B2L model uses a background layer and a foreground prompt. We compare our approach against all these three models. 
For LayerDiffuse, we use the official models based on Stable Diffusion (SD) 1.5~\cite{ldm}.
For the F2L and B2L models, we first generate foreground and background layers using SD 1.5, then use the generated layers as their input.

For comparison, we constructed a test set of 572 triplets of foreground, background, and composite prompts using ChatGPT to ensure diverse and robust evaluation scenarios. We generated layered images at a resolution of $768\times768$ using three models of LayerDiffuse~\cite{layerdiffuse} and \MethodName{}. The phrases used with ChatGPT for constructing the test set and examples of the generated triplets are included in the supplementary material.

\paragraph{Qualitative Comparisons}
\cref{fig:5_1_qualitative_comparison} provides a qualitative comparison against LayerDiffuse~\cite{layerdiffuse}. The LayerDiffuse models generate disproportionately large foregrounds that do not blend naturally with their backgrounds. This issue arises because they learn the foreground distribution from an RGBA dataset that focuses on individual objects and introduces unwanted bias into trained models. Additionally, LayerDiffuse struggles to handle actions such as hopping or barking, as it loses corresponding knowledge during fine-tuning. In contrast, our method generates high-quality layers that naturally blend and accurately fit the input prompt.

\cref{fig:5_1_qualitative_comparison2} presents a qualitative comparison of our method with Text2Layer~\cite{text2layer} and LayerDiff~\cite{layerdiff}. 
In the figure, Text2Layer produces low-quality textures and inaccurate masks due to their suboptimal training dataset synthesis approach. Similarly, LayerDiff produces foreground and background layers that do not blend naturally due to their approach of simultaneously synthesizing both layers from scratch and their binary mask-based layer representation. In contrast, our method generates high-quality foreground and background layers that blend seamlessly into naturally-composed composite images.

\paragraph{Quantitative Comparisons}
We perform quantitative evaluation using several metrics. To assess image quality, we estimate distribution similarity between synthesized layers and reference datasets using FID~\cite{FID} and KID~\cite{KID}. For the reference datasets, we use the MAGICK dataset~\cite{magick} for the foreground layer and the COCO dataset~\cite{cocodataset} for both the background layer and composite image. To evaluate text alignment for each layer, we use the CLIP score~\cite{clipscore}, which measures the cosine similarity between an image and a prompt in the CLIP embedding space.

Additionally, we propose a novel FG-MIoU score to assess the quality of both foreground and background layers. The FG-MIoU score is calculated as follows: for each layer, we first detect a foreground bounding box using a foreground prompt and GroundingDINO~\cite{groundingdino}. From the bounding box, we estimate a semantic mask using SAM~\cite{sam} and calculate the Mean Intersection over Union (MIoU) between the semantic mask and the layer.
The FG-MIoU evaluates different aspects for the foreground and background layers. For the foreground layer, it assesses the quality of the foreground shape by inspecting whether the generated foreground is accurately detected by recognition models. For the background layer, it evaluates the clean separation between layers by checking for any foreground objects detected within the background.

\cref{tab:tbl_1_main_quantitative_evaluation} reports the quantitative evaluation results. Overall, our method achieves superior scores compared to existing methods, owing to our layering strategy that leverages the generative power of existing models trained on large-scale datasets and ensures accurate layer decomposition. For FID and KID, our method shows slightly less favorable results for the foreground layer. We attribute this to the similarity between the MAGICK dataset and the training data of LayerDiffuse~\cite{layerdiffuse}, which focuses on foreground objects.

For a comprehensive evaluation of our approach, we also assess the diversity of the positions and scales of synthesized foreground objects. \cref{tab:tbl_2_foreground_layer_analysis} reports the means and standard deviations of four different metrics: the occupancy ratio, longest span ratio, and vertical and horizontal centers. The occupancy ratio indicates the proportion of image pixels occupied by foreground objects relative to the total pixel count. The longest span ratio measures the largest dimension of foreground objects, i.e., $\mathrm{max}(H, W)$, as a fraction of the corresponding axis length. Lastly, the vertical and horizontal centers represent the central positions of the foreground objects along each axis. The four metrics are normalized on a scale from 0 to 100.

As reported in \cref{tab:tbl_2_foreground_layer_analysis}, LayerDiffuse~\cite{layerdiffuse} generates large foreground objects that occupy substantial portions of images, with minimal variation in scale and position. Notably, the average longest span exceeds 90\%. In contrast, our method generates foreground layers with greater diversity in both scale and position, providing a more varied range of layered image synthesis.

\paragraph{User Study}
We conducted a user study to evaluate each method from a human perspective. 
To this end, 24 participants from our institution were recruited. 
Each participant reviewed 60 examples generated by three baseline methods and our method for 15 test prompts. For each example containing foreground, background, and final composite images, participants rated the quality of each image on a scale from 1 (poor) to 5 (excellent) based on two criteria: (a) alignment with the prompt text (Text Alignment) and (b) aesthetic quality and naturalness of the image (Image Quality).
\cref{tab:tbl_3_userstudy} presents the results, demonstrating that our method achieves the highest scores for all layers on both criteria.
The user study interface and questionnaires are included in the supplementary material.

\subsection{Evaluation of Layering Stage}
\begin{figure}
    \centering
    \includegraphics[width=\linewidth]{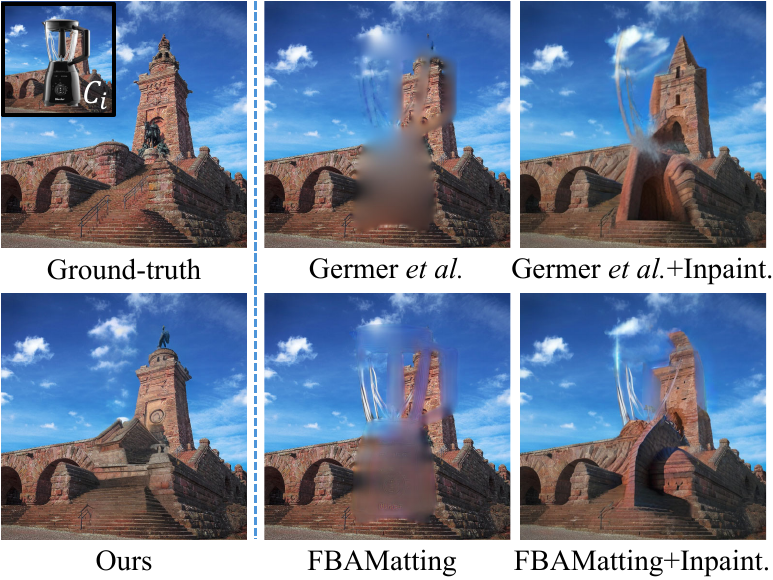}
    \vspace{-0.6cm}
    \caption{
    Qualitative comparison of background layers produced by the baseline methods and our method. 
    The inset in the top-left image represents input composite image.
    Germer~\etal~\cite{fastcolor} and FBAMatting~\cite{fbalphamatting} produce blurry background layers with artifacts due to the lack of generative capability.
    Evan with additional inpainting, they fail to produce natural background layers.
    Our method synthesizes an accurate and natural background layers by effectively leveraging visible information in input image.
    }
    \label{fig:5_2_compared_to_naive}
    \vspace{-0.6cm}
\end{figure}
\begin{table}[t]
\begin{center}
\caption{Quantitative evaluation of layer decomposition by na\"ive baselines and \MethodName{} using different metrics.
Here, MAD and MSE is presented multiplied by $10^3$, and LPIPS by $10^2$. SAD is presented divided by $10^{-3}$.}
\vspace{-0.2cm}
\label{tab:tbl_4_comparison_naive}
\scalebox{0.80}{
\begin{tabular}{c|c|cccc}

\Xhline{3\arrayrulewidth}
                       &                   & MAD$\downarrow$  & MSE$\downarrow$  & SAD$\downarrow$  & LPIPS$\downarrow$ \\
\hline

\multirow{5}{*}{FG}    & Germer~\etal~\cite{fastcolor}                     & \phantom{0}2.58                                   & \phantom{0}0.32                                   & \phantom{0}2.03                                     & \phantom{0}1.51 \\
                       & Germer~\etal~\cite{fastcolor} + Inp.              & \phantom{0}2.58                                   & \phantom{0}0.32                                   & \phantom{0}2.03                                     & \phantom{0}1.51 \\
                       & FBAMatting~\cite{fbalphamatting}                    & \phantom{0}3.54                                   & \phantom{0}0.94                                   & \phantom{0}2.78                                     & 2.52 \\
                       & FBAMatting~\cite{fbalphamatting} + Inp.             & \phantom{0}3.54                                   & \phantom{0}0.94                                   & \phantom{0}2.78                                     & 2.52 \\
                       & \cellcolor[HTML]{FFF9C4} Ours & \cellcolor[HTML]{FFF9C4}\phantom{0}\textbf{2.10}  & \cellcolor[HTML]{FFF9C4}\phantom{0}\textbf{0.31}  & \cellcolor[HTML]{FFF9C4}\phantom{0}\textbf{1.65}    & \cellcolor[HTML]{FFF9C4}\textbf{1.33} \\
\hline

\multirow{5}{*}{BG}    & Germer~\etal~\cite{fastcolor}                     & 48.83                                  & 12.76                                  & 38.40                                    & 28.73 \\
                       & Germer~\etal~\cite{fastcolor} + Inp.              & 57.22                                  & 19.74                                  & 45.00                                    & 26.87 \\
                       & FBAMatting~\cite{fbalphamatting}                    & \textbf{44.91}                         & \textbf{11.56}                         & \textbf{35.32}                           & 28.32 \\
                       & FBAMatting~\cite{fbalphamatting} + Inp.             & 57.34                                  & 20.42                                  & 45.10                                    &  25.90\\
                       & \cellcolor[HTML]{FFF9C4}Ours  & \cellcolor[HTML]{FFF9C4}55.52          & \cellcolor[HTML]{FFF9C4}17.52          & \cellcolor[HTML]{FFF9C4}43.66            & \cellcolor[HTML]{FFF9C4}\textbf{21.80} \\
\Xhline{3\arrayrulewidth}

\end{tabular}
}
\end{center}
\vspace{-0.9cm}
\end{table}
We assess the layering performance of the proposed layering stage by comparing it with four baselines that na\"ively combine image matting and inpainting.
From a composite image and a trimap, the first and second baselines estimate the foreground and background layers. The first baseline first estimates an alpha mask using ViTMatte~\cite{vitmatte} and then determines the colors of each layer using an optimization-based method of Germer~\etal~\cite{fastcolor}.
The second baseline directly estimates each layer using a network-based matting approach, FBAMatting~\cite{fbalphamatting}.
Since previous methods lack the capability to generate content, which is essential for handling large occluded regions, we additionally perform inpainting on the background estimated by the first and second baselines with a binary mask for areas where $\alpha>0.95$, and we set them as the third and fourth baselines.

For evaluation, we construct a test set of 1,000 synthetic composite images and trimaps for image matting. The composite images are synthesized by combining randomly-sampled foreground and background images from the test sets of the MAGICK~\cite{magick} and BG-20k~\cite{bg20k} datasets, respectively. 
The trimaps are synthesized following Xu~\etal~\cite{deepimagematting}.

\cref{fig:5_2_compared_to_naive} compares background layers produced by the baseline methods and ours. 
The baselines without additional inpainting, Germer~\etal's method~\cite{fastcolor} and FBAMatting~\cite{fbalphamatting}, produce blurry results with artifacts due to the lack of generative capability.
While additional inpainting produces more realistic results in the occluded regions, it cannot handle artifacts in regions where $0<\alpha<0.95$, producing unnatural background layers. 
In contrast, our layering stage effectively utilizes visible background areas to achieve an accurate and contextually coherent background decomposition.

We also report a quantitative assessment in \cref{tab:tbl_4_comparison_naive}.
For the foreground layer, our method achieves the most favorable scores across all metrics.
In terms of the background layer,
while our method achieves the best score in LPIPS,
Germer~\etal's method~\cite{fastcolor} and FBAMatting~\cite{fbalphamatting} outperform ours in MAD, MSE, and SAD. This is primarily because both approaches produce smooth results in occluded regions, which are expected to be closer to the ground truth than those with synthesized high-frequency details.
Compared to the other methods with additional inpainting, our method achieves better scores because it effectively utilizes information from visible background areas.

\subsection{Applications}
\begin{figure}
    \centering
    \includegraphics[width=\linewidth]{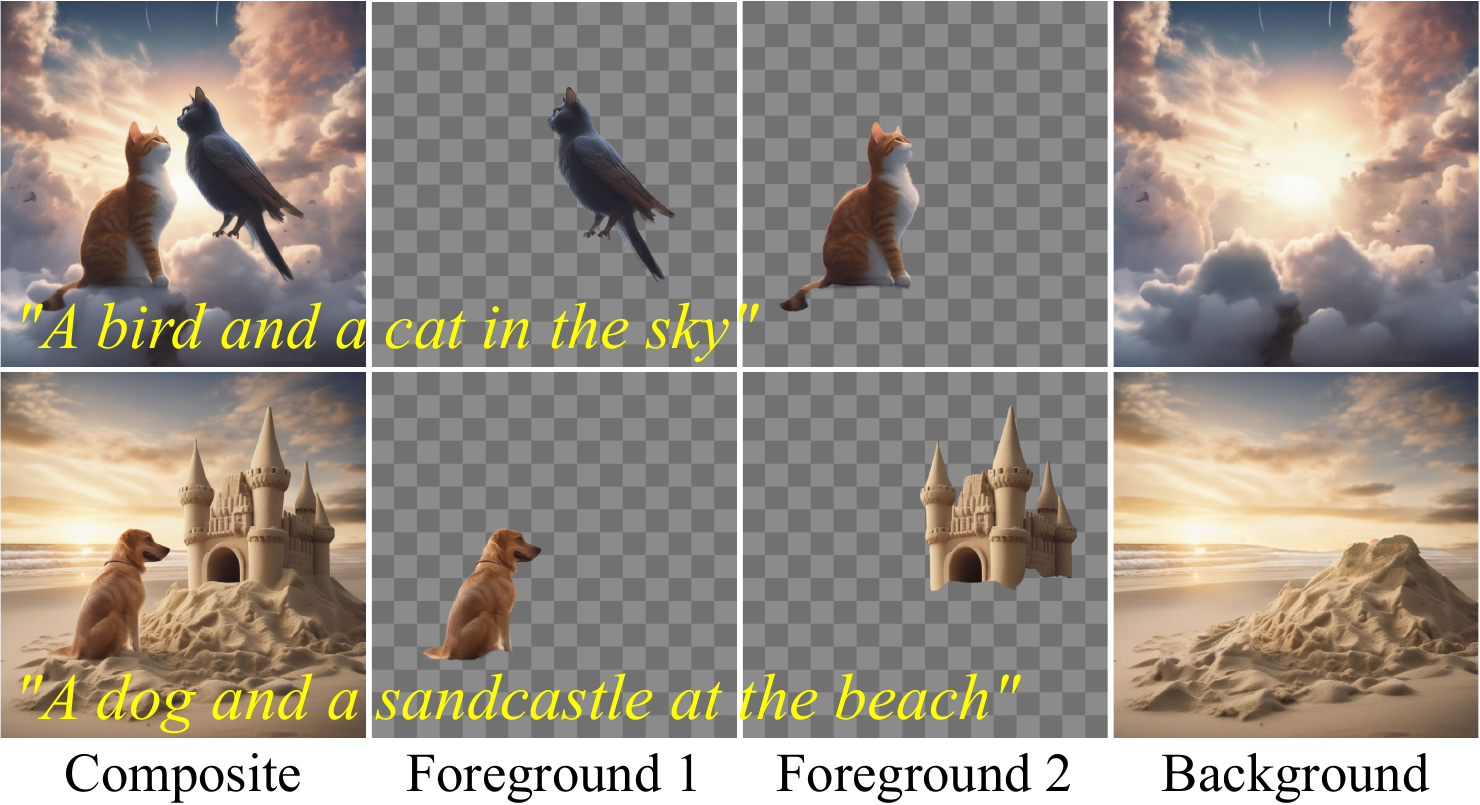}
    \vspace{-0.6cm}
    \caption{
    Application 1: multi-layered image synthesis. Starting from an initial composite image containing multiple foreground objects, \MethodName{} can generate a layered image composed of multiple layers through sequential inference. 
    }
    \label{fig:5_3_app1_multi_layer}
    \vspace{-0.2cm}
\end{figure}

\begin{figure}
    \centering
    \includegraphics[width=\linewidth]{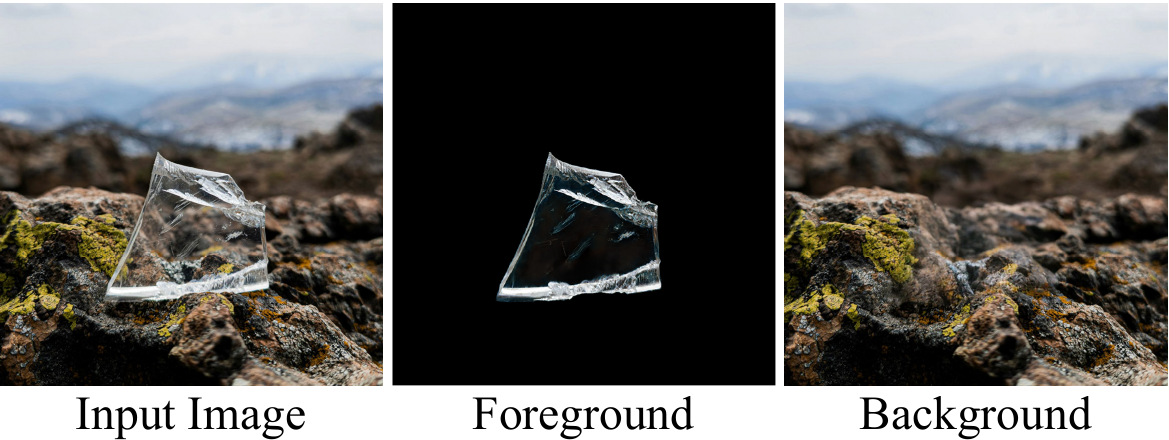}
    \vspace{-0.6cm}
    \caption{
    Application 2: layer decomposition on real-world image. Our layering stage successfully decompose real-world images, not limited in synthetic image decomposition.
    }
    \label{fig:5_3_app2_real_world}
    \vspace{-0.4cm}
\end{figure}

\MethodName{} can be easily extended for various applications and enhance their practicality. Here, we demonstrate notable applications of our approach.
We include \textbf{more applications and examples} in the supplementary material.

\paragraph{Multi-layered Image Synthesis}
\MethodName{} can also generate a layered image with multiple foreground layers through sequential decomposition.
As shown in \cref{fig:5_3_app1_multi_layer}, from the input prompts highlighted in yellow, composite images are generated and then decomposed into multiple layers.

\paragraph{Real-world Image Decomposition}
Our layering stage can also be used for layer decomposition of real-world images, as shown in \cref{fig:5_3_app2_real_world}, greatly broadening its applicability.

\section{Conclusion}
In this paper, we proposed \MethodName{}, an effective pipeline for synthesizing layered images from user prompts.
By decomposing an initial composite image into its constituent layers, \MethodName{} achieves high-quality layered image generation without large-scale training. For effective layer decomposition, we introduced adaptation of generative prior and a high-frequency alignment strategy.
Through extensive experiments, we demonstrate the effectiveness of our method and showcase its diverse applications.

\paragraph{Limitations.}
\MethodName{} is not free from limitations. 
It assumes an accurate alpha prediction, and inaccurate predictions can compromise layer quality. Additionally, if shadows are present in the initial composite images, shadows may remain in the background after layer decomposition, creating an unnatural effect and potentially requiring further shadow removal process. Detailed discussions and illustrative examples of these limitations are provided in the supplementary material.
{
    \small
    \bibliographystyle{ieeenat_fullname}
    \bibliography{main}
}


\end{document}